%% file: acl_latex.tex
\pdfoutput=1

\documentclass[11pt]{article}

\include{personal_definitions}

\usepackage[preprint]{acl}

\usepackage{times}
\usepackage{latexsym}
\usepackage{amsmath}
\usepackage{amsfonts}

\usepackage{enumitem}

\usepackage[T1]{fontenc}

\usepackage[utf8]{inputenc}

\usepackage{microtype}
\usepackage{algorithm}
\usepackage{algpseudocode}
\usepackage{inconsolata}

\usepackage{booktabs}

\usepackage{graphicx, caption, subcaption}
\usepackage{subcaption}

\title{On the Limited Generalization Capability of the Implicit Reward Model  Induced by Direct Preference Optimization}

\author{
    Yong Lin\thanks{Equal contribution}\thanks{Work done during internship at Apple}\thanks{The Hong Kong University of Science and Technology}
     Skyler Seto\footnotemark[1]\thanks{Apple} Maartje ter Hoeve\footnotemark[4] Katherine Metcalf\footnotemark[4] 
     \AND 
     Barry-John Theobald\footnotemark[4] Xuan Wang\footnotemark[4]
     Yizhe Zhang\footnotemark[4] Chen Huang\footnotemark[4]
     Tong Zhang\thanks{University of Illinois Urbana-Champaign}
}

\begin{document}
\maketitle
\begin{abstract}
Reinforcement Learning from Human Feedback (RLHF) is an effective approach for aligning language models to human preferences and reducing risks in deploying them in the wild. Central to RLHF is learning a reward function for scoring human preferences. Two main approaches for learning a reward model are 1) training an EXplicit Reward Model (EXRM) as in RLHF, and 2) using an implicit reward learned from preference data through methods such as Direct Preference Optimization (DPO).  Prior work has shown that the implicit reward model of DPO (denoted as DPORM) can approximate an EXRM in the limit. However, it is unclear how well DPORM empirically matches the performance of EXRM. DPORM's effectiveness directly implies the optimality of the learned policy, and also impacts preference labeling in LLM alignment methods including iterative DPO.  This work studies the accuracy at distinguishing preferred and rejected answers for both DPORM and EXRM.  Our findings indicate that even though DPORM fits the training dataset comparably, it generalizes less effectively than EXRM, especially when the validation datasets contain distribution shifts. Across five out-of-distribution settings, DPORM has a mean drop in accuracy of 3\% and a maximum drop of 7\%. These findings highlight that DPORM has limited generalization ability and substantiates the integration of an explicit reward model in iterative DPO approaches.

\end{abstract}

\input{sections/intro}

\input{sections/basics}

\input{sections/experiments}

\input{sections/conclusion}

\input{sections/limitations}

\section*{Acknowledgements}

We are grateful to Zak Aldeneh, Richard Bai, Dan Busbridge, Navdeep Jaitly, and Josh Susskind for their helpful discussions, comments, and
thoughtful feedback in reviewing this work.

\bibliography{custom}

\clearpage 

\appendix

\input{sections/appendix}

\end{document}

%% file: personal_definitions.tex
\usepackage{amsmath}
\usepackage{amssymb}
\DeclareMathOperator*{\argmin}{arg\,min}
\DeclareMathOperator*{\argmax}{arg\,max}

\def\ref{\mathrm{ref}} 
\def\rm{\mathrm{rm}}

\def\\avg{\avg}

%% file: sections/intro.tex
\section{Introduction}

Large language models (LLMs) have demonstrated exceptional performance on many tasks in diverse applications, including mathematical reasoning, coding capabilities, and knowledge-based question answering \citep{brown2020language,bubeck2023sparks, OpenAI2023GPT4TR}. Whilst LLMs  have broad knowledge and reasoning skills, the pre-training objective is often misaligned from the objective of instruction following \citep{ouyang2022training} according to human preferences \citep{christiano2017deep}, and LLMs can exhibit undesirable behaviors including hallucinating, and providing harmful or biased instructions \citep{huang2023survey, zhang2023r}. As LLMs become more commonplace, it is important for them to be aligned with human preferences for helpfulness, harmlessness, and honesty \citep{bai2022training}. 

\begin{figure}
        \centering
        \includegraphics[width=0.75\linewidth]{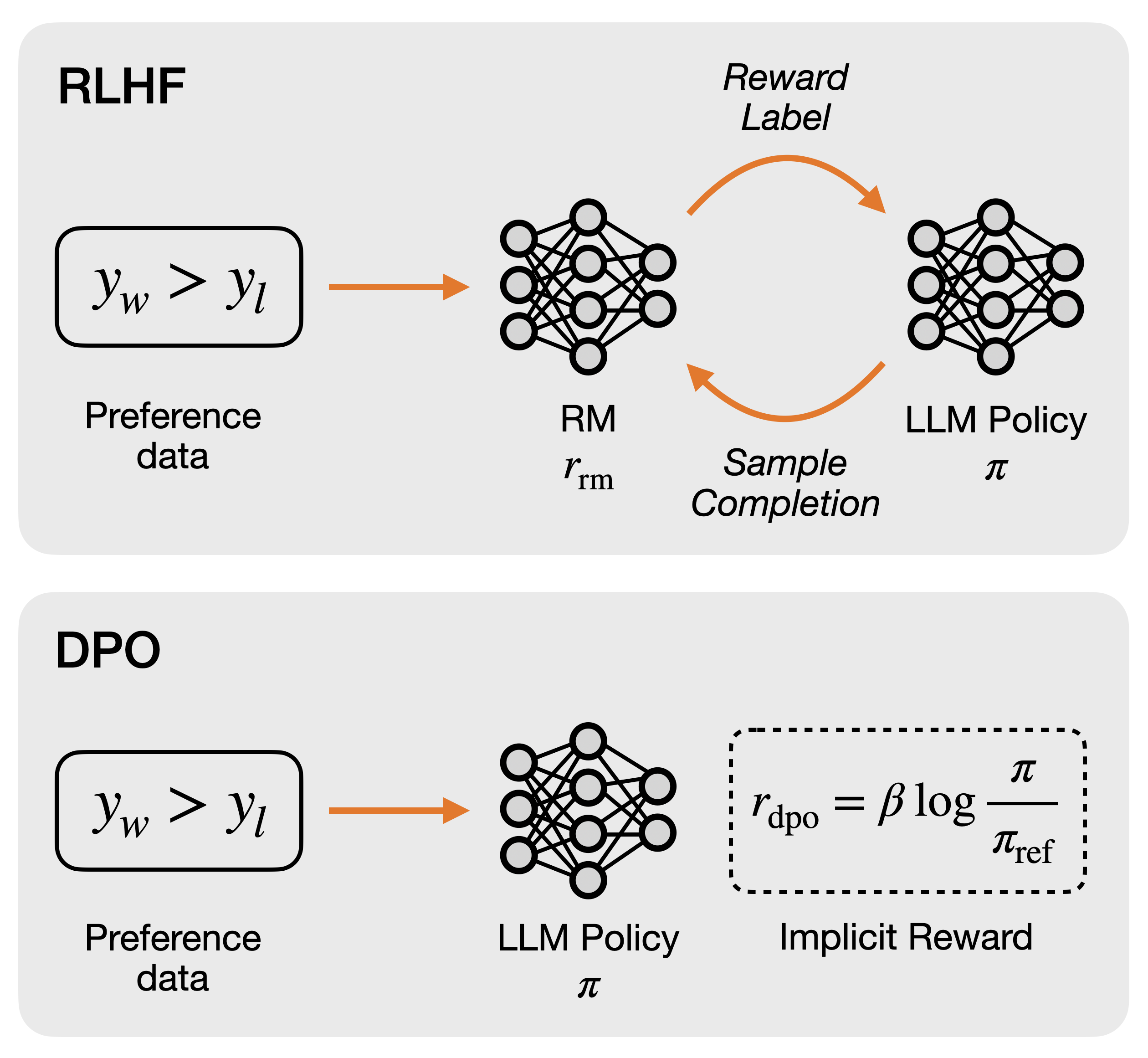}
        \caption{Overview of methods for learning reward models explicitly and implicitly (via DPO). Figure adapted from \citet{rafailov2024direct}.}
        \label{fig:pipline}
\end{figure}

A common practice for aligning LLMs to human preference is through Reinforcement Learning with Human Feedback (RLHF) \citep{ouyang2022training}, which is based on a reward model trained to score model outputs according to human preference annotations. %
In RLHF alignment, a high quality reward model is required for policy learning \citep{rame2024warm}. However, in practice, learned reward models are typically imperfect approximations of the ``true'' human reward label function \citep{gao2023scaling}, because they are trained on a fixed set of human preference data collected offline. When used within the RLHF policy optimization, they may see out-of-distribution (OOD) data when annotating the response of the model. Aligning according to an imperfect reward model can lead to worse performing language models as policy optimization continues to optimize a mis-specified reward. This can lead to an increased gap between the learned and true reward, a phenomena known as over-optimization and reward hacking \citep{gao2023scaling,skalse2022defining}.

\begin{figure}
    \centering
    \includegraphics[width=\linewidth]{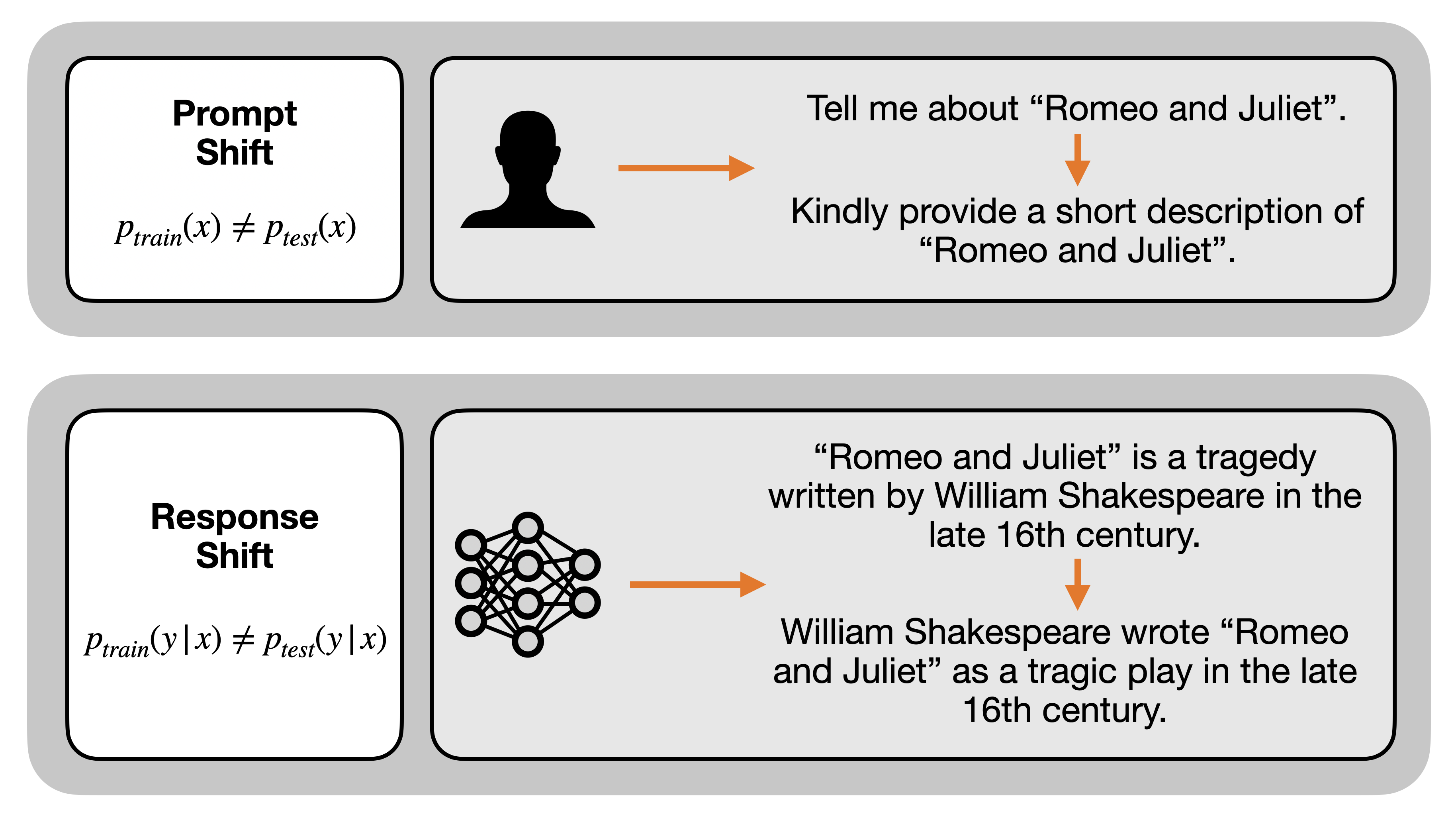}
    \caption{Examples of different types of distributional shifts for reward models and accuracy drops on real-world datasets.}
    \label{fig:dataset_shifts}
\end{figure}

Recently, \citet{rafailov2024direct} proposed Direct Preference Optimization (DPO), %
and show that any reward model can be implicitly represented by the optimal policy learned by DPO and a reference policy under certain assumptions. Similar to RLHF, DPO assumes access to a preference dataset, but directly finetunes a language model policy by minimizing a negative log likelihood objective. Due to its simplicity, DPO offers a more stable and convenient alignment process \citep{liu2023statistical}.

While, DPO is often regarded as simpler alternative to RLHF fine-tuning, and models trained with DPO have been shown to empirically match or outperform the RLHF policy performance even on OOD data \citep{rafailov2024direct}, its generalization ability still remains under-explored.

In this work, we conduct a systematic comparison between EXplicit Reward Models (EXRM) learned by RLHF and DPO's implicit Reward Model (DPORM) in terms of their generalization ability at distinguishing preferred and rejected answers. We train EXRM and DPORM on datasets containing chat, instruction following, and summarization, and evaluate them on in-distribution (ID)  evaluation sets, and ten OOD evaluation sets. Across five train-test shifts, and three model series: Gemma-2B, Gemma-7B \citep{team2024gemma} and Mistral-7B (with instruction tuning) \citep{jiang2023mistral} totaling $35$ experiments, we find that DPORM underperforms EXRM when the validation dataset contains distributional shifts. DPORM has a mean drop in accuracy of 3\% and a maximum drop of 7\%. Our results highlight the importance of learning an EXRM and justifies more complex online approaches that combine EXRM with iterative DPO \citep{xu2023some,liu2023statistical}.

%% file: sections/basics.tex
\section{Background} 

\begin{table*}[]
    \centering
    \begin{tabular}{c|cccc}
    \toprule
       \textbf{Settings}  & \textbf{Training Set} & \textbf{ID Testing} & \textbf{OOD Testing} & \textbf{Shift Type} \\
        \midrule
       Setting I(a)  & Ultra & Ultra  & Arena, HH, Nectar, WebGPT, Sum & Mixture\\
       Setting I(b)  & Arena & Arena & Ultra, HH, Nectar, WebGPT, Sum & Mixture\\
       Setting I(c)  & HH & HH & Ultra, Arena, Nectar, WebGPT, Sum& Mixture \\
       \midrule
       Setting II(a)& Ultra & Ultra&Ultra-LM3-SFT, Ultra-LM3-RLHF& Response shift\\
       Setting II(b)& Sum-Reddit & Sum-Reddit & Sum-DailyMail, Sum-CNN& Prompt shift\\
           \bottomrule
    \end{tabular}
    \caption{Detailed experimental settings. Note HH, Ultra and Sum are short for HH-RLHF, UltraFeedBack and Summarisation dataset, respectively.}
    \label{tab:settings}
\end{table*}

Training a reward model $r$ involves training a classifier according to a preference dataset $D$. The reward model takes as input a prompt $x \in \mathcal{X}$ and response $y \in \mathcal{Y}$ pair, and scores the response $r(x, y)$. The reward model is typically parameterized by a language model with an additional linear layer, the output of which is a scalar reward, which is used to compute the preference probability. An overview of the learning procedures RLHF via DPO and RL algorithms are included in Figure~\ref{fig:pipline}.

Given a set of collected preference datasets ${D} = \{x^{(i)}, y^{(i)}_w, y^{(i)}_{\ell}\}_{i=1}^{N}$ where $y^{(i)}_w$ and $y^{(i)}_{\ell}$ are the chosen and rejected responses to the prompt $x^{(i)}$, and $N$ is the number of samples, an explicit reward model $r(x, y)$ is trained by minimizing the negative log-likelihood over the preference dataset

\begin{align}
    \label{eqn:reward_estimation}
    r_{\mathrm{rm}}(x, y) &= \max_{\phi} \Big(-\mathbb{E}_{(x, y_w, y_l) \sim \mathcal{D}} \nonumber \\
    & [\log\sigma(r_\phi(x, y_w)  - r_\phi(x, y_l))] \Big).
\end{align}

RLHF methods train a LLM $\pi_{\mathrm{rm}}$ to maximize the reward given by $r_{\mathrm{rm}}$: 

\begin{align}
    \pi_{\rm} &= \max_{\pi_\theta} \mathbb{E}{x\sim \mathcal{D}, y\sim \pi(\cdot|x; \theta)}\big[r_{\rm}(x, y) \nonumber \\
    & -\beta KL(\pi(\cdot|x; \theta) || \pi_{\mathrm{ref}}(\cdot|x) ) \big].
    \label{eqn:reward_maximization}
\end{align}

Direct Preference Optimization \citep{rafailov2024direct} assumes there is a ground truth reward function $r^*$ and that the human preferences follow the Bradley-Terry (BT) model \citep{biomet1952}, such that given a prompt $x$ with two responses $y_w$ and $y_{\ell}$, the probability that $y_w$ is preferred over $y_{\ell}$ is:
\begin{align}
    \label{eqn:bt_model}
    \mathbb{P}(y_w \succ y_{\ell} | x) = \sigma(r^*(x, y_{w}) - r^*(x, y_{\ell})),
\end{align}
where $\sigma(z) = 1/(1 + \exp(-z)), \forall z \in \mathbb{R}$ and $y_w \succ y_{\ell}$ means the response $y_w$ is preferred over $y_{\ell}$.  

Under these assumptions, an aligned LLM $\pi_{\mathrm{dpo}}$ can be optimized without explicitly training a reward model. Combining Equation~\ref{eqn:reward_maximization} with \ref{eqn:bt_model}, the RL optimization can be reduced to the objective:
\begin{align}
    \pi_{\mathrm{dpo}}  &= \min_{\pi_\theta} \mathbb{E}_{(x, y_w, y_l) \sim \mathcal{D}} \nonumber \\
    &  \Bigg[\log\sigma \left(\beta \frac{\pi_{\theta}(y_w|x)}{\pi_{\mathrm{ref}}(y_w|x)} - \beta \frac{\pi_{\theta}(y_l|x)}{\pi_{\mathrm{ref}}(y_l|x)}\right) \Bigg],
    \label{enq:dpo_policy}
\end{align}
and the implicit reward model can be expressed in terms of the DPO policy as 

\begin{align}
    \label{eqn:dpo_rm}
    r_{\mathrm{dpo}}(x, y) = \beta \log \frac{\pi_{\mathrm{dpo}}(y|x)}{\pi_{\mathrm{ref}}(y|x)}. 
\end{align}
where $\pi_{\mathrm{ref}}(y|x)$ is a reference language model.  Because DPO parameterizes its reward function through the language model, the quality of the reward model is conditioned on the generative power of the language model \citep{li2024rl}.  Consequently, the implicit reward model can face issues when the representation features are mis-specified, for example due to different training datasets, or different architectures \citep{li2024rl,xu2024dpo}.  Prior work also conjectures that the task of generating preferred responses as in DPO's objective is more challenging than learning a discriminator between responses \citep{dong2024rlhf}.  
 In Section~\ref{sec:experiments}, we investigate the generalization capability of EXRM (Equation~\ref{eqn:reward_estimation}) and DPORM (Equation~\ref{eqn:dpo_rm}).

%% file: sections/experiments.tex
\section{Experiments}
\label{sec:experiments}

To study the generalization ability of EXRM and DPORM, we investigate both the ID performance on a held-out validation set, and the OOD generalization performance. We study two distribution shifts summarized in Figure~\ref{fig:dataset_shifts}: (1) \textbf{Prompt Shift}: the distribution of the testing prompts differs from that of training prompts. This shift could occur when training a reward model on prompts within some domain and using the reward model to annotate the prompts from  other domains as in iterative DPO. (2) \textbf{Response Shift}: the distributions of the  training and testing prompts are the same, whereas the responses to the prompts come from different distributions. This shift could occur when a different model is used to generate responses for a given prompt set, or in online updates of the model \citep{dong2024rlhf,liu2023statistical,meng2024simpo}. In the following, we conduct experiments in two settings: (I) evaluation of reward models on data from different sources, which contain a mixture of the distribution shifts introduced above (Section~\ref{sec:exp_set_1}), (II) controlled evaluations on  the two types of the distribution shifts (Section~\ref{sec:exp_set_2}) with  datasets and shift types enumerated in Table~\ref{tab:settings}. We additionally investigate the impact of reward model generalization capability on alignment by both EXRM and DPORM in an iterative DPO setting in Section~\ref{sec:iterative_dpo}.

For all experiments we finetune decoder-only transformer models \citep{vaswani2017attention} at 2B and 7B model parameter scales. The models are trained using the TRL\footnote{\url{https://github.com/huggingface/trl}} library. For reward modeling we fine-tune all models for 1 epoch using a learning rate of $5e^{-6}$.  For DPO, we train for two epochs using a learning rate of $1e^{-6}$ and $\beta=0.03$. All other hyperparameters correspond to the default parameter setting in the TRL library. Hyper-parameters for all model-data combinations were selected using Gemma-2B and 7B with UltraFeedBack as the training set.   Details of our sweep are given in Section~\ref{sec:hyperparams}.  For all experiments, unless otherwise stated we report the mean and standard deviations of the accuracy over three random seeds.

   \begin{figure*}[ht]
    \centering
    \hfill
       \begin{subfigure}{0.37\textwidth}
        \centering
        \includegraphics[width=\textwidth]{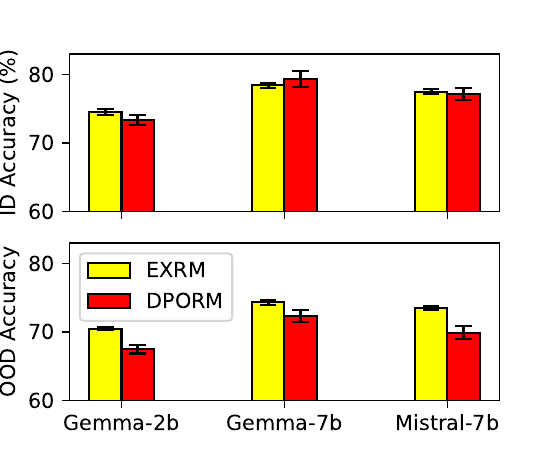}
        \caption{}
        \label{fig:all_shift}
        \vspace{-3mm}
    \end{subfigure}
    \hfill
    \begin{subfigure}{0.24\textwidth}
        \centering
        \includegraphics[width=\columnwidth]{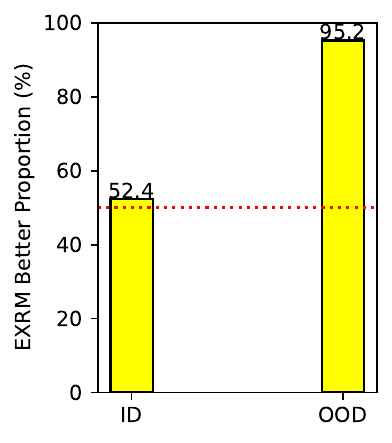}
         \caption{}
       \label{fig:detail_rm_dpo}
        \vspace{-2mm}

    \end{subfigure}
    \hfill
    \begin{subfigure}{0.23\textwidth}
        \centering
      \includegraphics[width=\textwidth]{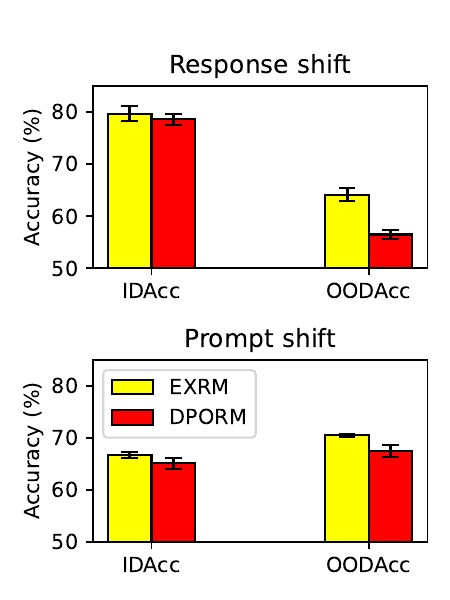}
        \caption{}

        \label{fig:shift_type}
        \vspace{-2mm}
    \end{subfigure}
    \caption{(a) The aggregated mean ID and OOD accuracy for different experiments across Setting I: a mixture of all distribution shifts in Table~\ref{tab:settings}. (b) The proportion of experiments where EXRM outperform DPORM in Setting I with three models and three seeds. (c) Results on specific types of distributional shift Setting II in Table~\ref{tab:settings}. (c-Top) The response shift evaluated on UltraFeedBack (ID) and our annotated dataset based on the generation of LLaMA3-8B (OOD). (c-Bottom) Prompt shift evaluated on summarization TL;DR (ID), CNN and DailyMail (OOD).}
    \label{fig:common_caption}
\end{figure*}

\subsection{Experimental Setting I: Mixture of Distributions}
\label{sec:exp_set_1}
We use six datasets: HH-RLHF \citep{bai2022training}, Arena, UltraFeedBack \citep{cui2023ultrafeedback}, Nectar \citep{starling2023}, Summarisation \citep{liu2020learning} and WebGPT \citep{nakano2021webgpt}. Dataset statistics are summarized in Table~\ref{tab:datasets_info} and described in greater detail in Appendix~\ref{app:datasets}.  
 We have three sub-settings I(a)-I(c) as shown in Table \ref{tab:settings}. In each setting, we train on one dataset (HH-RLHF, Arena or UltraFeedBack), and then evaluate on the respective ID validation split as well as 5 other datasets as OOD evaluation sets. For example, in Sub-setting I(a), we train on UltraFeedBack and use HH-RLHF, Arena, Nectar, Summarisation and WebGPT to evaluate OOD performance.

We use three instruction-tuned LLMs: Gemma-2B, Gemma-7B~\citep{team2024gemma}, and Mistral-7B~\citep{jiang2023mistral}. Figure~\ref{fig:common_caption}(a) aggregates the results of Settings I(a)-I(c) to report their average ID and OOD accuracy. We highlight the following observations: (1) DPORM and EXRM have a similar ID accuracy. This is evidenced in Figure~\ref{fig:common_caption}(b), which shows the proportion of experiments where EXRM outperforms DPORM. The results show that EXRM has higher ID accuracy than DPORM in 52\% of the total experiments indicating an equal win rate for the two reward models.   (2) While both DPORM and EXRM experience performance drops in OOD data on all three models, \emph{DPORM suffers from a larger drop and achieves consistently inferior OOD performance to EXRM} even on datasets where DPORM performed better on the ID data (additional details for individual evaluation sets and RewardBench benchmarks in Section~\ref{sec:detailed_results}). The win rate of EXRM on OOD data increases to over 90\% highlighting a lack of generalization capability for DPORM.

\subsection{Experimental Setting II: Controlled Distribution Shifts}
\label{sec:exp_set_2}
Based on results from Section~\ref{sec:exp_set_1}, we explore the impact of singular distribution shifts. In particular, we study (1) \textbf{Prompt Shift}, where we train on the Reddit TL;DR subset of Summarisation (denoted by Sum-Reddit) and evaluate on the CNN and DailyMail subsets (Denoted by Sum-CNN and DailyMail, respectively). These subsets are annotated by the same labelers in \citep{liu2020learning}, however the prompts in these subsets contain distribution shifts. (2) \textbf{Response Shift}, where we experiment with UltraFeedBack, and induce a response shift by generating responses using the SFT LLaMA3-8B-Instruct\footnote{\url{https://huggingface.co/meta-llama/Meta-Llama-3-8B-Instruct}} model and LLaMA3-8B model after iterative DPO\footnote{\url{https://huggingface.co/RLHFlow/LLaMA3-iterative-DPO-final}}, we generate responses for the validation set of Ultra. GPT4 is used to annotate pairwise response preferences following the protocol in creating Ultra \citep{cui2023ultrafeedback}. The resulting datasets are referred to as Ultra-LM3-SFT and Ultra-LM3-RLHF (Table~\ref{tab:settings}). We finetune the instruction-tuned Gemma-2B using the same hyper-parameters. The results  in Figure~\ref{fig:common_caption}(c) show that DPO consistently under-performs explicit reward modeling under both response and prompt shifts.

\subsection{Iterative Alignment Under Distribution Shifts}
\label{sec:iterative_dpo}
 Due to the potential limitations of the implicit reward model, recent works study combining DPO with an explicit reward model through iterative training \citep{xu2023some,xiong2024iterative}, and sample selection \citep{liu2023statistical}.  Still other work uses the implicit reward  of DPO and observe similar gains \citep{yuan2024self}. While the results summarized in Figure~\ref{fig:common_caption} show that DPORM underperforms EXRM under distribution shifts, the capability of DPORM towards aligning LLMs is still important to investigate.

To demonstrate the impact a reward model with worse generalization capability has on the alignment procedure, we fine-tune language models using Algorithm~\ref{algo:iter_DPO}, similar to \citep{dong2024rlhf,liu2023statistical,meng2024simpo}.  We study alignment with iterative DPO as it shows stronger performance than offline DPO, and has distribution shifts within the alignment process as both the model updates during training, and the prompt set for the iterative stage is different from those used to train the reward model \citep{dong2024rlhf}.

For iterative DPO experiments, we fine-tune the instruction-tuned Gemma-2B model. We first train the DPORM and EXRM using the UltraFeedback dataset and the later iterative procedure is conducted on the prompt set provided in RLHF Workflow\footnote{\url{https://github.com/RLHFlow/Online-RLHF}}. To compare the impact of EXRM and DPORM on alignment, we consider the above algorithm where the RM in Step 3 is either a separately trained EXRM on the original preference dataset or the DPORM from the policy trained with DPO. 

Results are reported in Table~\ref{tab:alignment_stage} on the AlpacaEval benchmark~\citep{li2023alpacaeval}, which measures the instruction following capability of language models by comparing responses generated by the model with GPT-4 \citep{OpenAI2023GPT4TR}. First, we find that the iterative DPO procedure improves with iterations over the base model with both DPORM and EXRM. However, the resulting win rate with EXRM exceeds the model trained with DPORM. This indicates that when the prompt set and model changes during training, the better robustness of the EXRM improves model training, confirming that our findings in Sections~\ref{sec:exp_set_1}-\ref{sec:exp_set_2} extend to instruction following capability of the final model.

\begin{table}[]
    \centering
    \begin{tabular}{c|cccc}
        \toprule
        \textbf{RM} & \textbf{Base} & \textbf{DPO}& \textbf{Iter 1} & \textbf{Iter 2}\\
        \midrule
       EXRM  & 5.88&10.09&13.56&13.86\\
       DPORM & 5.88&10.09&11.02&11.13\\
       \bottomrule
    \end{tabular}
     \caption{Length Controlled Win-rate (\%) over GPT-4 on Alpaca Eval for models trained with iterative DPO using an EXRM or DPORM. The base model is Gemma-2b instruction tuned model.}
    \label{tab:alignment_stage}
\end{table}

%% file: sections/conclusion.tex
\section{Conclusion and Discussion}
This work takes a step towards explicitly characterizing the generalization ability of the implicit reward function of DPO compared with an explicitly trained reward model. Our findings highlight that the implicit reward model consistently underperforms the explicit reward model. For researchers and practitioners aligning LLMs, our work sheds light on the potential benefits of using reinforcement learning algorithms to fine-tune over the simpler approach of DPO, and substantiates the field of recent work investigating iterative DPO algorithms that combine explicit reward models with DPO.

%% file: sections/limitations.tex
\section{Limitations}
\label{sec:limit}

This work focuses on training reward models (2B-7B) across a range of datasets.  However, the impact of %
the models themselves
is difficult to control as these models are pre-trained on data which is not publicly available.  Additionally, varying model sizes may be impacted more. %
We focused on 2B-7B models, as these are commonly benchmarked reward models \citep{lambert2024rewardbench}.

Finally, this work focuses on aligning language models to human preferences with primary applications including chat, summarization, and instruction following. Beyond such applications, there are a range of applications not covered by this work including code generation \citep{xu2023wizardlm}, and reasoning tasks \citep{chung2024scaling}, which are left to future work.  This limitation further extends to our focus on English in this work.
While there is significant interest in training English language models, it remains important future work to understand to what extent results hold for other languages.

%% file: sections/appendix.tex
\section{Additional Details for Out-of-Distribution Problems in RLHF}

\paragraph{Out-of-Distribution Problems in RLHF.} Understanding the impact of out-of-distribution (OOD) data on reward models is an ongoing and important direction.  The primary issue for training results from the offline three-step process of RLHF which trains the reward model on a static preference dataset.  When the reward model is subsequently optimized during LLM fine-tuning, the generated samples may appear out of distribution resulting in improper rewards and over optimization \cite{eisenstein2023reward,gao2023scaling,lang2024fine,lin2023speciality,yang2024regularizing}. Prior works studied properties of RLHF robustness including reward hacking \cite{skalse2022defining,pang2022reward}, and underspecification \cite{eisenstein2023helping,eisenstein2023reward}, but did not conduct systematic studies across different tasks, and types of reward models. Other benchmarks evaluate reward models and RLHF across a range of tasks, however they do not control for OOD robustness as they do not control the training data \cite{lambert2024rewardbench,dong2024rlhf}. Multiple prior works aim to enhance the reward model's OOD generalization ability by leveraging model ensembles \cite{coste2023reward,eisenstein2023helping,rame2024warm, zhang2024improving,lin2023spurious}, adversarial data augmentation \cite{cheng2023adversarial}, multiple attribute data annotation \cite{wang2024arithmetic} or on-policy data \cite{lang2024fine, zhai2024fine}. %

\section{Dataset Descriptions}
\label{app:datasets}

Details of the preferences and collection of each dataset are summarized below.  Table~\ref{tab:datasets_info} includes details about the response and annotation types as well as the dataset sizes.  Note that in our experiments we train and test on both a variety of dataset sizes, annotation schemes, and tasks.
\begin{itemize}[leftmargin=*]
   \item \textbf{HH-RLHF }\cite{bai2022training} contains multi-round conversations between users and Claude. The chosen and rejected answers are selected by humans.  
    \item \textbf{Chatbot Arena Conversations} \cite{zheng2023judging} contains 33K cleaned conversations with pairwise human preferences. It is collected on the Chatbot Arena from April to June 2023. Each sample includes the conversation between user and two models. 
    \item \textbf{UltraFeedBack} \cite{cui2023ultrafeedback} contains prompts from a wide range of datasets, i.e., UltraChat, FLAN, FalseQA, TruthfulQA, Evol-Instrauct, and ShareGPT. For each prompt, responses are generated by multiple, high-quality LLMs. GPT4 selects the chosen versus rejected answers.
    \item \textbf{UltraFeedBack-Binarized-Cleaned} is a subset of UltraFeedBack that does not include samples with prompts from Truthful-QA.
    \item \textbf{Summarisation} \cite{liu2020learning} is  a TL;DR dataset from Reddit posts on a variety of topics, as well summaries of the posts written by the users. The preference labels are provided by humans.
    \item \textbf{Nectar} \cite{starling2023}. Nectar's prompts are from  a diverse set of sources, including lmsys-chat-1M, ShareGPT, Antropic/hh-rlhf, UltraFeedback, Evol-Instruct, and Flan. They use  GPT-4, GPT-3.5-turbo, GPT-3.5-turbo-instruct, LLama-2-7B-chat, and Mistral-7B-Instruct to generate responses and GPT-4 to select the chosen versus rejected responses.
    \item \textbf{WebGPT} \cite{nakano2021webgpt} contains prompts  from the “Explain Like I’m Five” subreddit. They collected answers generated by human and models with the web-browsing environment. Humans select the chosen versus rejected responses.
\end{itemize}

\begin{table}[t!]
    \centering
    \begin{tabular}{c|cccc}
        \toprule
        \textbf{Dataset} & \textbf{Size} & \textbf{Resp.} & \textbf{Ann.}  \\
        \midrule
        HH-RLHF & 115K & LLM& Human\\
        UltraFeedBack& 340K& LLM& GPT4\\
        Nectar& 365K& LLM& GPT4\\
        Arena & 22K &  LLM & Human\\
        WebGPT& 13K &LLM & Human\\
        Summarisation& 92K & LLM & Human\\
        \bottomrule
    \end{tabular}
    \caption{Dataset statistics. Note the responses of different datasets are derived from different LLMs as discussed in Appendix~\ref{app:datasets}. Resp. = Response, Ann. = Annotator.}
    \label{tab:datasets_info}
\end{table}

\section{Model Descriptions}
\label{app:datasets}

Details of the language models we finetune are detailed.  We use three finetuned LLMs: 
\begin{itemize}[leftmargin=*]
    \item \textbf{Gemma-2B}: \url{https://huggingface.co/google/gemma-2b-it},
    \item  \textbf{Gemma-7B}: \url{https://huggingface.co/google/gemma-7b-it},
    \item \textbf{Mistral-7B}: \url{https://huggingface.co/HuggingFaceH4/mistral-7b-sft-beta}. 
\end{itemize}

\section{Experimental Details}
\label{sec:hyperparams}
We conduct our experiments based on Huggingface's TRL\footnote{\url{https://github.com/huggingface/trl}} package. We conduct a grid search for the hyper-parameters search with Gemma 2B and 7B (instruction-tuned) models on UltraFeedBack and use the hyper-parameter found with best ID valuation accuracy for other experimental settings. For reward modeling, we sweep learning rates in $[1e^{-5}, 5e^{-6}, 1e^{-6}]$ and epochs in $[1, 2, 3]$ following \cite{tunstall2023zephyr}. The best hyper-parameters are learning rate$ = 5e^{-6}$ and  epoch = $1$.  Reward model accuracy over all hyperparameters are provided for the Gemma-7B EXRM model in Table~\ref{tab:dporm_hp}

For DPO, we sweep learning rates in $[5e^{-6}, 1e^{-6}, 5e^{-7}]$, $\beta$ in $[0.01, 0.03, 0.1]$, epochs in $[1, 2, 3]$. The best hyper-parameter we found are learning rate$ = 1e^{-6}$, $\beta=0.03$ and  epoch = $2$. For all other hyper-parameters, we adopt the default setting in TRL.  Reward model accuracy over all hyperparameters are provided for the Gemma-7B DPORM model in Table~\ref{tab:dporm_hp}. Results for the 2B models follow similar patterns.

\begin{table}[t]
\centering
\begin{tabular}{ccc}
\toprule
\textbf{Epoch} & \textbf{LR}   & \textbf{Val Acc (\%)} \\
\midrule
1     & 1e-6 & 81.1 \\
\textbf{1}     & \textbf{5e-6} & \textbf{81.4} \\
1     & 1e-5 & 80.8 \\
2     & 1e-6 & 81.2 \\
2     & 5e-6 & 80.9 \\
2     & 1e-5 & 80.0 \\
\bottomrule
\end{tabular}
\caption{Validation Accuracy for Different Epochs and Learning Rates (in \%)}
\label{tab:exrm_hp}
\end{table}

\begin{table}[h!]
\centering
\begin{tabular}{cccc}
\toprule
\textbf{Epoch} & \textbf{Beta} & \textbf{LR}   & \textbf{Val Acc (\%)} \\
\midrule
1     & 0.03 & 1e-6 & 78.1 \\
1     & 0.03 & 5e-6 & 80.0 \\
1     & 0.1  & 1e-6 & 79.1 \\
1     & 0.1  & 5e-6 & 80.0 \\
\textbf{2}     & \textbf{0.03} & \textbf{1e-6} & \textbf{80.5} \\
2     & 0.03 & 5e-6 & 80.2 \\
2     & 0.1  & 1e-6 & 79.2 \\
2     & 0.1  & 5e-6 & 79.9 \\
3     & 0.03 & 1e-6 & 79.7 \\
3     & 0.03 & 5e-6 & 79.1 \\
3     & 0.1  & 1e-6 & 80.4 \\
3     & 0.1  & 5e-6 & 78.7 \\
\bottomrule
\end{tabular}
\caption{Validation Accuracy for Different Epochs, Beta Values, and Learning Rates}
\label{tab:dporm_hp}
\end{table}

Training a 2B reward model on the Ultrafeedback dataset ($\sim 100k$ samples) takes $\sim 12$ GPUh using  A100 GPUs, while a 7B model takes 36 GPUh.  DPO models take roughly twice as long as they are trained for two epochs.  

\section{Iterative DPO Algorithm}
We conduct experiments with iterative DPO using the RLHFlow library\footnote{\url{https://github.com/RLHFlow/Online-RLHF}}.  We use the default hyperparameters setting $K=8$, learning rate $5e-7$ with cosine learning rate scheduler, max steps $1200$, and max-min for chosen preferences.  The iterative DPO algorithm is provided in Algorithm~\ref{algo:iter_DPO}.

\begin{algorithm}[ht!]
    \caption{Iterative DPO}
    \label{algo:iter_DPO}
    \begin{algorithmic}[1] %
            \State \textbf{Input}: prompt set $\mathcal{S} = \{x^{(i)}\}_{i=1}^M$,  preference dataset $\mathcal{D}=\{x^{(i)}, y_w^{(i)}, y_l^{(i)}\}_{i=1}^N$, initial policy $\pi$, and sample size per prompt $K$.
            \State Obtain DPO policy $\pi_{0}$ from $\pi$ by \eqref{enq:dpo_policy}. 
            \State Obtain $r$ through \eqref{eqn:reward_estimation} or \eqref{eqn:dpo_rm}.
            \For{Iteration $t=1\dots T$}
                \State Set $\mathcal{D}_{t}$ as the empty set $\{\}$. 
                \For{Prompt $x^{(j)}$ in $\mathcal{S}$}.
                \State Sample response $y_1^{(j)}, ..., y_K^{(j)} \sim \pi_{\mathrm{iter} {(i-1)}}(\cdot|x^{(j)})$.
                \State Annotate $r_k^{(j)} = r(x^{(j)}, y^{(j)}_k)$ for $k=1, ..., K$.
                \State Select the chosen sample $y^{(j)}_{\bar k}$ and rejected sample $y^{(j)}_{\underline k}$where $ \bar k 
                = \argmax_k r_k $ and $ \underline k 
                = \argmin_k r_k $.
                \State $\mathcal{D}_{t} \xleftarrow[]{} \mathcal{D}_{t} \cup \{(x^{(j)}, y_{\bar k}^{(j)}, y_{\underline k}^{(j)}  )\}$.
                \EndFor
                \State Obtain $\pi_{t}$ by \eqref{enq:dpo_policy} with $\mathcal{D}_{t}$. 
            \EndFor
            
            \State \textbf{Output} $\pi_{0} ... \pi_{T}$.
    \end{algorithmic}
\end{algorithm}

\section{Results of Section I}
\label{sec:detailed_results}

\subsection{Detailed Results for Setting (I)}
Table~\ref{tab:detailed_result} shows the ID and average OOD performance for each model and training datasets, expanding on the results in Figure~\ref{fig:all_shift} and \ref{fig:detail_rm_dpo}. We see that even in settings where DPO performs better (2\%) in ID settings such as Mistral-7B trained on Arena and Gemma-7B trained on Arena and HH, the OOD performance drops by 1-2\%, and  in settings such as Gemma-2B and Mistral-7B trained on HH, where the ID accuracy is similar, the OOD accuracy decreases by 4-5\%. 

\begin{table*}[!htb]
    \centering
\begin{tabular}{cllll}
\toprule
\textbf{Model} & \textbf{Training Set} & \textbf{Method} &  \textbf{ID Acc (\%)} & \textbf{OOD Acc (\%)} \\
\midrule
 {Gemma-2b} & {Arena} & DPORM & 74.6 $\pm$ 1.0 & 59.2 $\pm$ 0.3 \\
 &  & EXRM & 75.3 $\pm$ 0.6 & 62.2 $\pm$ 0.2 \\
\midrule
 & {HH} & DPORM & 70.8 $\pm$ 0.7 & 59.7 $\pm$ 0.7 \\
 &  & EXRM & 70.8 $\pm$ 0.5 & 64.1 $\pm$ 0.4 \\
\midrule
 & {Ultra} & DPORM & 74.8 $\pm$ 0.5 & 62.0 $\pm$ 0.9 \\
 &  & EXRM & 77.5 $\pm$ 0.3 & 65.6 $\pm$ 0.2 \\
\midrule
{Gemma-7b} & {Arena} & DPORM & 81.2 $\pm$ 3.7 & 64.3 $\pm$ 1.4 \\
 &  & EXRM & 79.1 $\pm$ 3.3 & 65.3 $\pm$ 3.1 \\
\midrule
 & {HH} & DPORM & 75.3 $\pm$ 1.7 & 66.3 $\pm$ 1.0 \\
 &  & EXRM & 71.7 $\pm$ 0.2 & 68.3 $\pm$ 0.5 \\
\midrule
 & {Ultra} & DPORM & 79.9 $\pm$ 0.7 & 67.7 $\pm$ 0.3 \\
 &  & EXRM & 82.5 $\pm$ 0.4 & 69.0 $\pm$ 0.4 \\
\midrule
{Mistral-7B} & {Arena} & DPORM & 81.1 $\pm$ 1.2 & 62.6 $\pm$ 0.7 \\
 &  & EXRM & 78.8 $\pm$ 0.5 & 64.6 $\pm$ 0.3 \\
\midrule
 & {HH} & DPORM & 70.7 $\pm$ 2.2 & 63.2 $\pm$ 1.5 \\
 &  & EXRM & 72.3 $\pm$ 0.1 & 68.1 $\pm$ 0.1 \\
\midrule
 &  {Ultra} & DPORM & 81.7 $\pm$ 0.6 & 66.4 $\pm$ 1.0 \\
 &  & EXRM & 81.6 $\pm$ 0.4 & 69.2 $\pm$ 0.3 \\
\bottomrule
\end{tabular}
\caption{ID and OOD accuracy for different train sets in Setting I in Table~\ref{tab:settings}.}
    \label{tab:detailed_result}
\end{table*}

\subsection{Detailed Results for Setting (I) for Individual Eval Sets}

We further compare the OOD accuracy per evaluation set in Table~\ref{tab:detailed_result_indiv} for the Gemma-2B model.  While the findings remain  consistent with Table~\ref{tab:detailed_result}, we note that it is not strictly the case that DPORM always underperforms EXRM on all datasets.  For example training on Arena and evaluating on Ultra or Nectar (OOD datasets) would result in a 1\% increase.  However, in contrast for heavy distribution shifts such as training on HH and evaluating on Nectar or Arena where the response and prompts have changed, we note substantial drop in performance for DPORM. In contrast, evaluating training on Arena and evaluating on Nectar - two chat datasets, resulted in improved performance from DPORM over EXRM.

\begin{table*}[!htb]
    \centering
    \begin{tabular}{clllllll}
        \toprule
        \textbf{Training Set}  & \textbf{Method} &  \textbf{Ultra} & \textbf{HH} & \textbf{WebGPT} & \textbf{Sum} & \textbf{Nectar} & \textbf{Arena}  \\
        \midrule
        Arena & DPORM &  67.02 & 52.39 & 58.64 & 55.19 & 68.75 & 72.87 \\
        & EXRM & 66.24 & 54.17 & 59.08 & 58.73 & 67.78 & 74.13 \\
        \midrule 
        HH & DPORM & 60.11 & 70.75 & 59.57 & 56.52 & 63.96 & 62.63 \\
         & EXRM & 63.92 & 70.42 & 59.50 & 55.22 & 68.93 & 67.35 \\
        \midrule 
        Ultra & DPORM & 74.20 & 56.78 & 57.31 & 60.64 & 65.43 & 71.14 \\
        & EXRM & 77.73 & 57.98 & 62.17 & 62.47 & 69.70 & 74.38\\
        \bottomrule 
    \end{tabular}
    \caption{ID and OOD accuracy for different train sets in Setting I in Table~\ref{tab:settings}.}
    \label{tab:detailed_result_indiv}
\end{table*}

\subsection{RewardBench Results}
Finally, we compare the performance of DPORM and EXRM trained on different preference datasets on the RewardBench, a collection of evaluation datasets spanning chat, reasoning, and safety for challenging OOD evaluations \cite{lambert2024rewardbench}.  We focus only on evaluation in comparison with the prior experiments as the datasets are small containing fewer than 1500 samples. 

Results for the Gemma-2B model are summarized in Table~\ref{tab:rewardbench}.  For the Arena dataset, we find surprisingly that DPORM outperforms EXRM on average, but training with HH and Ultra leads to EXRM performing better.  We conjecture that this may be due to RewardBench having two datasets with chat.  However, we also note that the reasoning performance is higher for DPORM when trained on Arena but lower for both HH and Ultra.  Understanding OOD reasoning capabilities from RLHF requires further investigation as our experiments do not test reasoning capabilities for ID data.

\begin{table*}[!htb]
    \centering
    \begin{tabular}{lllllll}
        \toprule
        \textbf{Training Set}  & \textbf{Method} &  \textbf{Chat Easy} & \textbf{Chat Hard} & \textbf{Safety} & \textbf{Reasoning} & \textbf{Avg} \\    
        \midrule
        Arena & DPORM & 95.83 & 37.5 & 33.33 & 73.64 & 60.08\\
         & EXRM & 89.61 & 35.86 & 37.70 & 62.02 &  56.30\\ \midrule
         HH & DPORM & 65.63 & 47.66 & 68.23 & 64.95 & 61.62\\
         & EXRM & 89.33 & 40.35 & 74.05 & 70.34 & 68.52\\ \midrule
         Ultra & DPORM & 93.75 & 39.06 & 50.52 & 71.74 & 63.77\\
         & EXRM & 95.79 & 46.60 & 52.43 & 82.95 & 69.44\\
        \bottomrule
    \end{tabular}
        \caption{RewardBench accuracy for DPORM and EXRM Gemma-2B models trained on Arena, HH, and UltraFeedback.}
    \label{tab:rewardbench}
\end{table*}

%% file: acl_latex.bbl
\begin{thebibliography}{45}
\providecommand{\natexlab}[1]{#1}

\bibitem[{Bai et~al.(2022)Bai, Jones, Ndousse, Askell, Chen, DasSarma, Drain, Fort, Ganguli, Henighan et~al.}]{bai2022training}
Yuntao Bai, Andy Jones, Kamal Ndousse, Amanda Askell, Anna Chen, Nova DasSarma, Dawn Drain, Stanislav Fort, Deep Ganguli, Tom Henighan, et~al. 2022.
\newblock Training a helpful and harmless assistant with reinforcement learning from human feedback.
\newblock \emph{arXiv preprint arXiv:2204.05862}.

\bibitem[{Bradley and Terry(1952)}]{biomet1952}
Ralph~Allan Bradley and Milton~E. Terry. 1952.
\newblock {Rank Analysis of Incomplete Block Designs: The Method of Paired Comparisons}.
\newblock \emph{Biometrika}, 39(3-4):324--345.

\bibitem[{Brown et~al.(2020)Brown, Mann, Ryder, Subbiah, Kaplan, Dhariwal, Neelakantan, Shyam, Sastry, Askell et~al.}]{brown2020language}
Tom Brown, Benjamin Mann, Nick Ryder, Melanie Subbiah, Jared~D Kaplan, Prafulla Dhariwal, Arvind Neelakantan, Pranav Shyam, Girish Sastry, Amanda Askell, et~al. 2020.
\newblock Language models are few-shot learners.
\newblock \emph{Advances in neural information processing systems}, 33:1877--1901.

\bibitem[{Bubeck et~al.(2023)Bubeck, Chandrasekaran, Eldan, Gehrke, Horvitz, Kamar, Lee, Lee, Li, Lundberg et~al.}]{bubeck2023sparks}
S{\'e}bastien Bubeck, Varun Chandrasekaran, Ronen Eldan, Johannes Gehrke, Eric Horvitz, Ece Kamar, Peter Lee, Yin~Tat Lee, Yuanzhi Li, Scott Lundberg, et~al. 2023.
\newblock Sparks of artificial general intelligence: Early experiments with gpt-4.
\newblock \emph{arXiv preprint arXiv:2303.12712}.

\bibitem[{Cheng et~al.(2023)Cheng, Yang, Li, Dai, and Du}]{cheng2023adversarial}
Pengyu Cheng, Yifan Yang, Jian Li, Yong Dai, and Nan Du. 2023.
\newblock Adversarial preference optimization.
\newblock \emph{arXiv preprint arXiv:2311.08045}.

\bibitem[{Christiano et~al.(2017)Christiano, Leike, Brown, Martic, Legg, and Amodei}]{christiano2017deep}
Paul~F Christiano, Jan Leike, Tom Brown, Miljan Martic, Shane Legg, and Dario Amodei. 2017.
\newblock Deep reinforcement learning from human preferences.
\newblock \emph{Advances in neural information processing systems}, 30.

\bibitem[{Chung et~al.(2024)Chung, Hou, Longpre, Zoph, Tay, Fedus, Li, Wang, Dehghani, Brahma et~al.}]{chung2024scaling}
Hyung~Won Chung, Le~Hou, Shayne Longpre, Barret Zoph, Yi~Tay, William Fedus, Yunxuan Li, Xuezhi Wang, Mostafa Dehghani, Siddhartha Brahma, et~al. 2024.
\newblock Scaling instruction-finetuned language models.
\newblock \emph{Journal of Machine Learning Research}, 25(70):1--53.

\bibitem[{Coste et~al.(2023)Coste, Anwar, Kirk, and Krueger}]{coste2023reward}
Thomas Coste, Usman Anwar, Robert Kirk, and David Krueger. 2023.
\newblock Reward model ensembles help mitigate overoptimization.
\newblock \emph{arXiv preprint arXiv:2310.02743}.

\bibitem[{Cui et~al.(2023)Cui, Yuan, Ding, Yao, Zhu, Ni, Xie, Liu, and Sun}]{cui2023ultrafeedback}
Ganqu Cui, Lifan Yuan, Ning Ding, Guanming Yao, Wei Zhu, Yuan Ni, Guotong Xie, Zhiyuan Liu, and Maosong Sun. 2023.
\newblock Ultrafeedback: Boosting language models with high-quality feedback.
\newblock \emph{arXiv preprint arXiv:2310.01377}.

\bibitem[{Dong et~al.(2024)Dong, Xiong, Pang, Wang, Zhao, Zhou, Jiang, Sahoo, Xiong, and Zhang}]{dong2024rlhf}
Hanze Dong, Wei Xiong, Bo~Pang, Haoxiang Wang, Han Zhao, Yingbo Zhou, Nan Jiang, Doyen Sahoo, Caiming Xiong, and Tong Zhang. 2024.
\newblock Rlhf workflow: From reward modeling to online rlhf.
\newblock \emph{arXiv preprint arXiv:2405.07863}.

\bibitem[{Eisenstein et~al.(2023{\natexlab{a}})Eisenstein, Berant, Nagpal, Agarwal, Beirami, D'Amour, Dvijotham, Heller, Pfohl, and Ramachandran}]{eisenstein2023reward}
Jacob Eisenstein, Jonathan Berant, Chirag Nagpal, Alekh Agarwal, Ahmad Beirami, Alexander~Nicholas D'Amour, Krishnamurthy~Dj Dvijotham, Katherine~A Heller, Stephen~Robert Pfohl, and Deepak Ramachandran. 2023{\natexlab{a}}.
\newblock Reward model underspecification in language model alignment.
\newblock In \emph{NeurIPS 2023 Workshop on Distribution Shifts: New Frontiers with Foundation Models}.

\bibitem[{Eisenstein et~al.(2023{\natexlab{b}})Eisenstein, Nagpal, Agarwal, Beirami, D'Amour, Dvijotham, Fisch, Heller, Pfohl, Ramachandran et~al.}]{eisenstein2023helping}
Jacob Eisenstein, Chirag Nagpal, Alekh Agarwal, Ahmad Beirami, Alex D'Amour, DJ~Dvijotham, Adam Fisch, Katherine Heller, Stephen Pfohl, Deepak Ramachandran, et~al. 2023{\natexlab{b}}.
\newblock Helping or herding? reward model ensembles mitigate but do not eliminate reward hacking.
\newblock \emph{arXiv preprint arXiv:2312.09244}.

\bibitem[{Gao et~al.(2023)Gao, Schulman, and Hilton}]{gao2023scaling}
Leo Gao, John Schulman, and Jacob Hilton. 2023.
\newblock Scaling laws for reward model overoptimization.
\newblock In \emph{International Conference on Machine Learning}, pages 10835--10866. PMLR.

\bibitem[{Huang et~al.(2023)Huang, Yu, Ma, Zhong, Feng, Wang, Chen, Peng, Feng, Qin et~al.}]{huang2023survey}
Lei Huang, Weijiang Yu, Weitao Ma, Weihong Zhong, Zhangyin Feng, Haotian Wang, Qianglong Chen, Weihua Peng, Xiaocheng Feng, Bing Qin, et~al. 2023.
\newblock A survey on hallucination in large language models: Principles, taxonomy, challenges, and open questions.
\newblock \emph{arXiv preprint arXiv:2311.05232}.

\bibitem[{Jiang et~al.(2023)Jiang, Sablayrolles, Mensch, Bamford, Chaplot, Casas, Bressand, Lengyel, Lample, Saulnier et~al.}]{jiang2023mistral}
Albert~Q Jiang, Alexandre Sablayrolles, Arthur Mensch, Chris Bamford, Devendra~Singh Chaplot, Diego de~las Casas, Florian Bressand, Gianna Lengyel, Guillaume Lample, Lucile Saulnier, et~al. 2023.
\newblock Mistral 7b.
\newblock \emph{arXiv preprint arXiv:2310.06825}.

\bibitem[{Lambert et~al.(2024)Lambert, Pyatkin, Morrison, Miranda, Lin, Chandu, Dziri, Kumar, Zick, Choi et~al.}]{lambert2024rewardbench}
Nathan Lambert, Valentina Pyatkin, Jacob Morrison, LJ~Miranda, Bill~Yuchen Lin, Khyathi Chandu, Nouha Dziri, Sachin Kumar, Tom Zick, Yejin Choi, et~al. 2024.
\newblock Rewardbench: Evaluating reward models for language modeling.
\newblock \emph{arXiv preprint arXiv:2403.13787}.

\bibitem[{Lang et~al.(2024)Lang, Huang, and Li}]{lang2024fine}
Hao Lang, Fei Huang, and Yongbin Li. 2024.
\newblock Fine-tuning language models with reward learning on policy.
\newblock \emph{arXiv preprint arXiv:2403.19279}.

\bibitem[{Li et~al.(2023)Li, Zhang, Dubois, Taori, Gulrajani, Guestrin, Liang, and Hashimoto}]{li2023alpacaeval}
Xuechen Li, Tianyi Zhang, Yann Dubois, Rohan Taori, Ishaan Gulrajani, Carlos Guestrin, Percy Liang, and Tatsunori~B Hashimoto. 2023.
\newblock Alpacaeval: An automatic evaluator of instruction-following models.

\bibitem[{Li et~al.(1972)Li, Xu, and Yu}]{li2024rl}
Ziniu Li, Tian Xu, and Yang Yu. 1972.
\newblock When is rl better than dpo in rlhf? a representation and optimization perspective.
\newblock In \emph{The Second Tiny Papers Track at ICLR 2024}.

\bibitem[{Lin et~al.(2023{\natexlab{a}})Lin, Tan, Hao, Wong, Dong, Zhang, Yang, and Zhang}]{lin2023spurious}
Yong Lin, Lu~Tan, Yifan Hao, Honam Wong, Hanze Dong, Weizhong Zhang, Yujiu Yang, and Tong Zhang. 2023{\natexlab{a}}.
\newblock Spurious feature diversification improves out-of-distribution generalization.
\newblock \emph{arXiv preprint arXiv:2309.17230}.

\bibitem[{Lin et~al.(2023{\natexlab{b}})Lin, Tan, Lin, Zheng, Pi, Zhang, Diao, Wang, Zhao, Yao et~al.}]{lin2023speciality}
Yong Lin, Lu~Tan, Hangyu Lin, Zeming Zheng, Renjie Pi, Jipeng Zhang, Shizhe Diao, Haoxiang Wang, Han Zhao, Yuan Yao, et~al. 2023{\natexlab{b}}.
\newblock Speciality vs generality: An empirical study on catastrophic forgetting in fine-tuning foundation models.
\newblock \emph{arXiv preprint arXiv:2309.06256}.

\bibitem[{Liu et~al.(2020)}]{liu2020learning}
Fei Liu et~al. 2020.
\newblock Learning to summarize from human feedback.
\newblock In \emph{Proceedings of the 58th Annual Meeting of the Association for Computational Linguistics}.

\bibitem[{Liu et~al.(2023)Liu, Zhao, Joshi, Khalman, Saleh, Liu, and Liu}]{liu2023statistical}
Tianqi Liu, Yao Zhao, Rishabh Joshi, Misha Khalman, Mohammad Saleh, Peter~J Liu, and Jialu Liu. 2023.
\newblock Statistical rejection sampling improves preference optimization.
\newblock \emph{arXiv preprint arXiv:2309.06657}.

\bibitem[{Meng et~al.(2024)Meng, Xia, and Chen}]{meng2024simpo}
Yu~Meng, Mengzhou Xia, and Danqi Chen. 2024.
\newblock Simpo: Simple preference optimization with a reference-free reward.
\newblock \emph{arXiv preprint arXiv:2405.14734}.

\bibitem[{Nakano et~al.(2021)Nakano, Hilton, Balaji, Wu, Ouyang, Kim, Hesse, Jain, Kosaraju, Saunders et~al.}]{nakano2021webgpt}
Reiichiro Nakano, Jacob Hilton, Suchir Balaji, Jeff Wu, Long Ouyang, Christina Kim, Christopher Hesse, Shantanu Jain, Vineet Kosaraju, William Saunders, et~al. 2021.
\newblock Webgpt: Browser-assisted question-answering with human feedback.
\newblock \emph{arXiv preprint arXiv:2112.09332}.

\bibitem[{OpenAI(2023)}]{OpenAI2023GPT4TR}
OpenAI. 2023.
\newblock Gpt-4 technical report.
\newblock \emph{ArXiv}, abs/2303.08774.

\bibitem[{Ouyang et~al.(2022)Ouyang, Wu, Jiang, Almeida, Wainwright, Mishkin, Zhang, Agarwal, Slama, Ray et~al.}]{ouyang2022training}
Long Ouyang, Jeffrey Wu, Xu~Jiang, Diogo Almeida, Carroll Wainwright, Pamela Mishkin, Chong Zhang, Sandhini Agarwal, Katarina Slama, Alex Ray, et~al. 2022.
\newblock Training language models to follow instructions with human feedback.
\newblock \emph{Advances in neural information processing systems}, 35:27730--27744.

\bibitem[{Pang et~al.(2022)Pang, Padmakumar, Sellam, Parikh, and He}]{pang2022reward}
Richard~Yuanzhe Pang, Vishakh Padmakumar, Thibault Sellam, Ankur~P Parikh, and He~He. 2022.
\newblock Reward gaming in conditional text generation.
\newblock \emph{arXiv preprint arXiv:2211.08714}.

\bibitem[{Rafailov et~al.(2024)Rafailov, Sharma, Mitchell, Manning, Ermon, and Finn}]{rafailov2024direct}
Rafael Rafailov, Archit Sharma, Eric Mitchell, Christopher~D Manning, Stefano Ermon, and Chelsea Finn. 2024.
\newblock Direct preference optimization: Your language model is secretly a reward model.
\newblock \emph{Advances in Neural Information Processing Systems}, 36.

\bibitem[{Ram{\'e} et~al.(2024)Ram{\'e}, Vieillard, Hussenot, Dadashi, Cideron, Bachem, and Ferret}]{rame2024warm}
Alexandre Ram{\'e}, Nino Vieillard, L{\'e}onard Hussenot, Robert Dadashi, Geoffrey Cideron, Olivier Bachem, and Johan Ferret. 2024.
\newblock Warm: On the benefits of weight averaged reward models.
\newblock \emph{arXiv preprint arXiv:2401.12187}.

\bibitem[{Skalse et~al.(2022)Skalse, Howe, Krasheninnikov, and Krueger}]{skalse2022defining}
Joar Skalse, Nikolaus Howe, Dmitrii Krasheninnikov, and David Krueger. 2022.
\newblock Defining and characterizing reward gaming.
\newblock \emph{Advances in Neural Information Processing Systems}, 35:9460--9471.

\bibitem[{Team et~al.(2024)Team, Mesnard, Hardin, Dadashi, Bhupatiraju, Pathak, Sifre, Rivi{\`e}re, Kale, Love et~al.}]{team2024gemma}
Gemma Team, Thomas Mesnard, Cassidy Hardin, Robert Dadashi, Surya Bhupatiraju, Shreya Pathak, Laurent Sifre, Morgane Rivi{\`e}re, Mihir~Sanjay Kale, Juliette Love, et~al. 2024.
\newblock Gemma: Open models based on gemini research and technology.
\newblock \emph{arXiv preprint arXiv:2403.08295}.

\bibitem[{Vaswani et~al.(2017)Vaswani, Shazeer, Parmar, Uszkoreit, Jones, Gomez, Kaiser, and Polosukhin}]{vaswani2017attention}
Ashish Vaswani, Noam Shazeer, Niki Parmar, Jakob Uszkoreit, Llion Jones, Aidan~N Gomez, \L~ukasz Kaiser, and Illia Polosukhin. 2017.
\newblock \href {https://proceedings.neurips.cc/paper_files/paper/2017/file/3f5ee243547dee91fbd053c1c4a845aa-Paper.pdf} {Attention is all you need}.
\newblock In \emph{Advances in Neural Information Processing Systems}, volume~30. Curran Associates, Inc.

\bibitem[{Wang et~al.(2024)Wang, Lin, Xiong, Yang, Diao, Qiu, Zhao, and Zhang}]{wang2024arithmetic}
Haoxiang Wang, Yong Lin, Wei Xiong, Rui Yang, Shizhe Diao, Shuang Qiu, Han Zhao, and Tong Zhang. 2024.
\newblock Arithmetic control of llms for diverse user preferences: Directional preference alignment with multi-objective rewards.
\newblock \emph{arXiv preprint arXiv:2402.18571}.

\bibitem[{Xiong et~al.(2024)Xiong, Dong, Ye, Wang, Zhong, Ji, Jiang, and Zhang}]{xiong2024iterative}
Wei Xiong, Hanze Dong, Chenlu Ye, Ziqi Wang, Han Zhong, Heng Ji, Nan Jiang, and Tong Zhang. 2024.
\newblock Iterative preference learning from human feedback: Bridging theory and practice for rlhf under kl-constraint.
\newblock In \emph{Forty-first International Conference on Machine Learning}.

\bibitem[{Xu et~al.(2023{\natexlab{a}})Xu, Sun, Zheng, Geng, Zhao, Feng, Tao, and Jiang}]{xu2023wizardlm}
Can Xu, Qingfeng Sun, Kai Zheng, Xiubo Geng, Pu~Zhao, Jiazhan Feng, Chongyang Tao, and Daxin Jiang. 2023{\natexlab{a}}.
\newblock Wizardlm: Empowering large language models to follow complex instructions.
\newblock \emph{arXiv preprint arXiv:2304.12244}.

\bibitem[{Xu et~al.(2023{\natexlab{b}})Xu, Lee, Sukhbaatar, and Weston}]{xu2023some}
Jing Xu, Andrew Lee, Sainbayar Sukhbaatar, and Jason Weston. 2023{\natexlab{b}}.
\newblock Some things are more cringe than others: Preference optimization with the pairwise cringe loss.
\newblock \emph{arXiv preprint arXiv:2312.16682}.

\bibitem[{Xu et~al.(2024)Xu, Fu, Gao, Ye, Liu, Mei, Wang, Yu, and Wu}]{xu2024dpo}
Shusheng Xu, Wei Fu, Jiaxuan Gao, Wenjie Ye, Weilin Liu, Zhiyu Mei, Guangju Wang, Chao Yu, and Yi~Wu. 2024.
\newblock Is dpo superior to ppo for llm alignment? a comprehensive study.
\newblock \emph{arXiv preprint arXiv:2404.10719}.

\bibitem[{Yang et~al.(2024)Yang, Ding, Lin, Zhang, and Zhang}]{yang2024regularizing}
Rui Yang, Ruomeng Ding, Yong Lin, Huan Zhang, and Tong Zhang. 2024.
\newblock Regularizing hidden states enables learning generalizable reward model for llms.
\newblock \emph{arXiv preprint arXiv:2406.10216}.

\bibitem[{Yuan et~al.(2024)Yuan, Pang, Cho, Sukhbaatar, Xu, and Weston}]{yuan2024self}
Weizhe Yuan, Richard~Yuanzhe Pang, Kyunghyun Cho, Sainbayar Sukhbaatar, Jing Xu, and Jason Weston. 2024.
\newblock Self-rewarding language models.
\newblock \emph{arXiv preprint arXiv:2401.10020}.

\bibitem[{Zhai et~al.(2024)Zhai, Bai, Lin, Pan, Tong, Zhou, Suhr, Xie, LeCun, Ma et~al.}]{zhai2024fine}
Yuexiang Zhai, Hao Bai, Zipeng Lin, Jiayi Pan, Shengbang Tong, Yifei Zhou, Alane Suhr, Saining Xie, Yann LeCun, Yi~Ma, et~al. 2024.
\newblock Fine-tuning large vision-language models as decision-making agents via reinforcement learning.
\newblock \emph{arXiv preprint arXiv:2405.10292}.

\bibitem[{Zhang et~al.(2023)Zhang, Diao, Lin, Fung, Lian, Wang, Chen, Ji, and Zhang}]{zhang2023r}
Hanning Zhang, Shizhe Diao, Yong Lin, Yi~R Fung, Qing Lian, Xingyao Wang, Yangyi Chen, Heng Ji, and Tong Zhang. 2023.
\newblock R-tuning: Teaching large language models to refuse unknown questions.
\newblock \emph{arXiv preprint arXiv:2311.09677}.

\bibitem[{Zhang et~al.(2024)Zhang, Chen, Chen, Shen, Sun, and Gan}]{zhang2024improving}
Shun Zhang, Zhenfang Chen, Sunli Chen, Yikang Shen, Zhiqing Sun, and Chuang Gan. 2024.
\newblock Improving reinforcement learning from human feedback with efficient reward model ensemble.
\newblock \emph{arXiv preprint arXiv:2401.16635}.

\bibitem[{Zheng et~al.(2023)Zheng, Chiang, Sheng, Zhuang, Wu, Zhuang, Lin, Li, Li, Xing, Zhang, Gonzalez, and Stoica}]{zheng2023judging}
Lianmin Zheng, Wei-Lin Chiang, Ying Sheng, Siyuan Zhuang, Zhanghao Wu, Yonghao Zhuang, Zi~Lin, Zhuohan Li, Dacheng Li, Eric.~P Xing, Hao Zhang, Joseph~E. Gonzalez, and Ion Stoica. 2023.
\newblock \href {https://arxiv.org/abs/2306.05685} {Judging llm-as-a-judge with mt-bench and chatbot arena}.
\newblock \emph{Preprint}, arXiv:2306.05685.

\bibitem[{Zhu et~al.(2023)Zhu, Frick, Wu, Zhu, and Jiao}]{starling2023}
Banghua Zhu, Evan Frick, Tianhao Wu, Hanlin Zhu, and Jiantao Jiao. 2023.
\newblock Starling-7b: Improving llm helpfulness and harmlessness with rlaif.

\end{thebibliography}
